\documentclass[conference]{IEEEtran}
\usepackage{times}
\usepackage{graphicx}
\usepackage{epsfig}
\usepackage{comment}
\usepackage{adjustbox}
\usepackage{amsmath}
\usepackage{color}

\usepackage{times}
\usepackage{epsfig}
\usepackage{graphicx}
\usepackage{amsmath}
\usepackage{amssymb}

\usepackage[numbers]{natbib}
\usepackage{multicol}
\usepackage[bookmarks=true]{hyperref}

\begin{document}

\title{Synthesizing Training Data for Object Detection in Indoor Scenes} 




%
\author{\authorblockN{Georgios Georgakis\authorrefmark{1},
Arsalan Mousavian\authorrefmark{1},
Alexander C. Berg\authorrefmark{2}, 
Jana Ko\v{s}eck{\'a}\authorrefmark{1} 
}
\authorblockA{\authorrefmark{1}George Mason University\\
\{ggeorgak,amousavi,kosecka\}@gmu.edu}
\authorblockA{\authorrefmark{2}University of North Carolina at Chapel Hill\\
aberg@cs.unc.edu}
}

\maketitle

\newcommand*{\arsalan}[1]{\textsf{\textcolor{red}{[{\bf Arsalan}: #1]}}}
\newcommand*{\george}[1]{\textsf{\textcolor{green}{[{\bf George}: #1]}}}
\newcommand*{\jana}[1]{\textsf{\textcolor{orange}{[{\bf Jana}: #1]}}}
\newcommand*\rot{\rotatebox{90}}

\begin{abstract}
Detection of objects in cluttered indoor environments is one of the key enabling functionalities for service robots. The best performing  object detection approaches in computer vision exploit deep Convolutional Neural Networks (CNN) to simultaneously detect and categorize the objects of interest in cluttered scenes. Training of such models typically requires large amounts of annotated training data which is time consuming and costly to obtain. In this work we explore the ability of using synthetically generated composite images for training state-of-the-art object detectors, especially for object instance detection. We superimpose 2D images of textured object models into images of real environments at variety of locations and scales. 
Our experiments evaluate different superimposition strategies ranging from purely image-based blending all the way to depth and semantics informed positioning of the object models into real scenes. 
We demonstrate the effectiveness of these object detector training strategies on two publicly available datasets, the GMU-Kitchens~\cite{Georgakis_3DV16} and the Washington RGB-D Scenes v2~\cite{Lai_icra14}. As one observation, augmenting some hand-labeled training data with synthetic examples carefully composed onto scenes yields object detectors with comparable performance to using much more hand-labeled data.
Broadly, this work charts new opportunities for training detectors for new objects by exploiting existing object model repositories in either a purely automatic fashion or with only a very small number of human-annotated examples. 


\end{abstract}

\IEEEpeerreviewmaketitle

\section{Introduction}
The capability of detecting and searching for common household objects in indoor environments is the key component  of the `fetch-and-delivery' task  commonly considered one of the main functionalities of service robots. Existing approaches for object detection are dominated by machine learning techniques focusing on learning suitable representations of object instances. This is especially the case when the objects of interest are to  be localized in environments with large amounts of clutter, variations in lighting, and a range of poses. While the problem of detecting object instances in simpler table top settings has been tackled previously using local features, these methods are often not effective in the presence of large amounts of clutter or when the scale of the objects is small.


Current leading object detectors exploit convolutional neural networks (CNNs) and are either trained end-to-end~\cite{Liu_ECCV16} for sliding-window detection or follow the region proposal approach which is jointly fine-tuned for accurate detection and classification~\cite{Girshick_ICCV15}~\cite{Ren_NIPS15}. In both approaches, the training and evaluation of object detectors requires labeling of a large number of training images with objects in various backgrounds and poses with the bounding boxes or even segmentations of objects from background.

\begin{figure}[t] 
      \centering
      \includegraphics[width=0.47\textwidth]{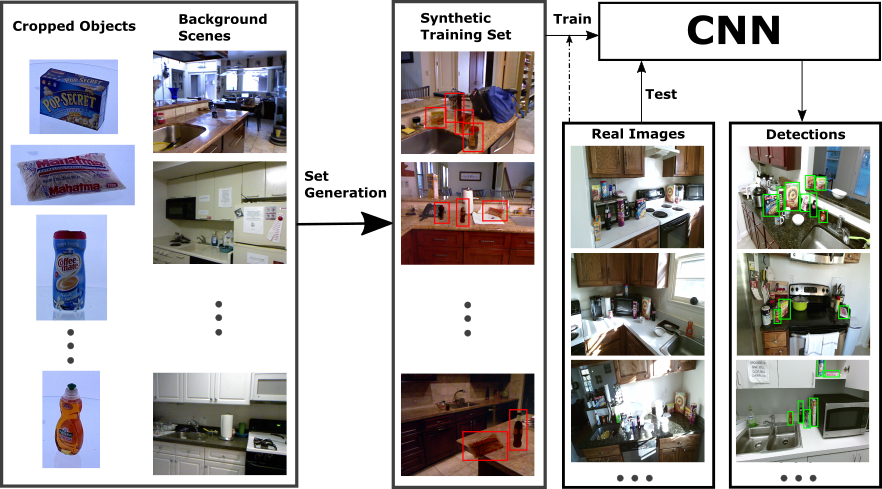}
      \caption{Given cropped object images and background scenes we propose an automated approach for generating synthetic training sets that can be used to train current state-of-the-art object detectors, which can then be applied to real test images. The generation procedure takes advantage of scene understanding methods in order to place the objects in meaningful positions in the images. We also explore using a combination of synthetic and real images for training and demonstrate higher detection accuracy compared to training with only real data. Best viewed in color.}
      \label{fig:title_image}
\end{figure}
Often in robotics, object detection is a prerequisite for tasks such as pose estimation, grasping, and manipulation. Notable efforts have been made to collect 3D models for object instances with and without textures, assuming that objects of interest are in proximity, typically on a table top.
Existing approaches to these challenges often use either 3D CAD models~\cite{DexNet_ICRA16} or texture mapped models of object instances obtained using traditional reconstruction pipelines ~\cite{Singh_ICRA14,Sun_WICRA16}. 

In this work we explore the feasibility of using such existing datasets of standalone objects on uniform backgrounds for training object detectors~\cite{Liu_ECCV16,Ren_NIPS15} that can be applied in real-world cluttered scenes. We create ``synthetic'' training images by superimposing the objects into images of real scenes. We investigate effects of different superimposition strategies ranging from purely image-based blending all the way to using depth and semantics to inform positioning of the objects. Toward this end we exploit the geometry and the semantic segmentation of a scene obtained using the state of the art method of~\cite{Arsalan_3DV16} to restrict the locations and size of the superimposed object model. We demonstrate that, in the context of robotics applications in indoor environments, these positioning strategies improve the final performance of the detector. This is in contrast with previous approaches~\cite{SaenkoICCV15,Su_ICCV15} which used large synthetic datasets with mostly randomized placement. 
In summary, our contributions are the following:
\begin{enumerate}
\item We propose an automated approach to generate synthetic training data for the task of object detection, which takes into consideration the geometry and semantic information of the scene.
\item Based on our results and observations, we offer insights regarding the superimposition design choices, that could potentially affect the way training sets for object detection are generated in the future.
\item We provide an extensive evaluation of current state-of-the-art object detectors and demonstrate their behavior under different training regimes.
\end{enumerate}




\section{Related Work}

We first briefly review related works in object detection to motivate our choice of detectors, then discuss previous attempts to use synthetic data as well as different datasets and evaluation methodologies.

\paragraph{Object Detection}
Traditional methods for object detection in cluttered scenes follow the sliding window based pipeline with hand designed flat feature representations (e.g. HOG) along with discriminative classifiers, such as linear or latent SVMs.  Examples include DPMs~\cite{Felzenszwalb_TPAMI10} which exploit efficient methods for feature computation and classifier evaluation. These models have been used successfully in robotics for detection in the table top setting~\cite{Lai_ICRA11}. Other effectively used strategies for object detection used local features and correspondences between a model reference image and the scene. These approaches~\cite{collet_IJRR11,Tang_ICRA12} worked well with textured household objects, taking advantage of the discriminative nature of the local descriptors. In an attempt to reduce the search space of the sliding window techniques, alternative approaches concentrated on generating category-independent object proposals~\cite{Uijlings_IJCV13,Cheng_CVPR14} using bottom up segmenation techniques followed by classification using traditional features. The flat engineered features have been recently superseded by 
approaches based on Convolutional Neural Networks (CNN's), which learn features with increased amount of invariance by repeated layering of convolutional and pooling layers. While these methods have been intially introduced for image classification task~\cite{Krizhevsky_NIPS12}, extensions to object detection include~\cite{Girshick_CVPR14}~\cite{SermanetCVPR13}. 
The R-CNN approach~\cite{Girshick_CVPR14} relied on finding object proposals and extracting features from each crop using a pre-trained network, making the proposal generating module independent from the classification module. Recent state 
of the art object detectors such as Faster R-CNN~\cite{Ren_NIPS15} and SSD~\cite{Liu_ECCV16} are trained jointly in a so called end-to-end fashion to both find object proposals and also classify them.

\paragraph{Synthetic Data} 
There are several previous attempts to use synthetic data for training CNNs. The work of~\cite{SaenkoICCV15} used existing 3D CAD models, both with and without texture, to generate 2D images by varying the projections and orientations of the objects. The approach was evaluated on 20 categories in PASCAL VOC2007 dataset. 
That work used earlier CNN models~\cite{Girshick_CVPR14} where the proposal generation module was independent from fine-tuning the CNN classifier, hence making the dependence on the context and background less prominent than in current models. In the work of~\cite{Su_ICCV15} the authors used the rendered models and their 2D projections on varying backgrounds to train a deep CNN for pose estimation. In these representative works, objects typically appeared on simpler backgrounds and were combined with the object detection strategies that rely on the proposal generation stage. Our work differs in that we perform informed compositing on the background scenes, instead of placing object-centric synthetic images at random locations. This allows us to train the CNN object detectors to produce higher quality object proposals, rather than relying on unsupervised bottom-up techniques. 
In~\cite{KoltunECCV16}, a Grand Theft Auto video game engine was used to collect scenes with realistic appearance and their associated category pixel level labels for the problem of semantic segmentation. Authors showed that using these high realism renderings can significantly reduce  the effort for annotation. They used a combination of synthetic data and real images to train models for semantic segmentation. Perhaps the closest work to ours is~\cite{Gupta_CVPR16}, which also generates a synthetic training set by taking advantage of scene segmentation to create synthetic training examples, however the task is that of text localization instead of object detection.

\begin{figure*}[t] 
      \centering
      \includegraphics[width=1\textwidth]{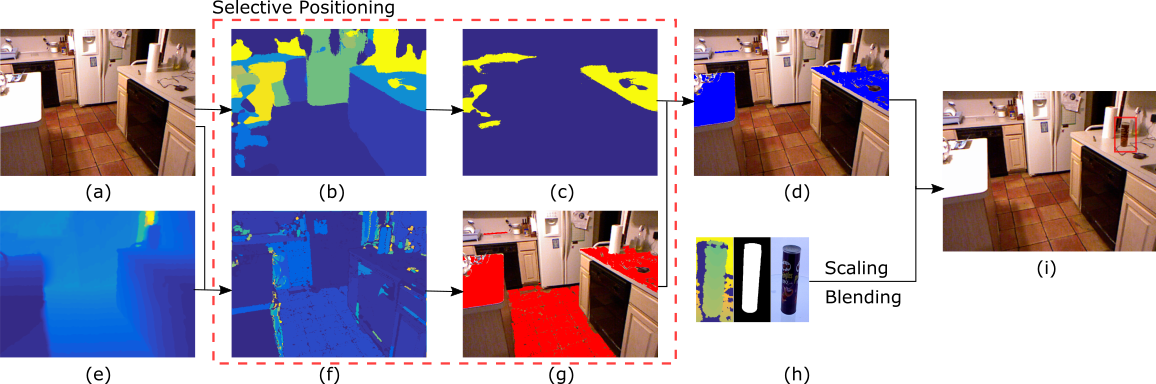}
      \caption{Overview of the procedure for blending an object in a background scene. We take advantage of estimated support surfaces (g) and predictions for counters and tables (c) in order to find regions for object placement (d). The semantic segmentation of the scene~\cite{Arsalan_3DV16}, and the plane extraction are shown in (b) and (f) respectively. (h) presents an example of an object's RGB, depth, and mask images, while (i) shows the final blending result. RGB and depth images of the background scene are in (a) and (e) respectively. Best viewed in color. }
      \label{fig:blending_procedure}
\end{figure*}
\section{Approach}

\subsection{Synthetic Set Generation}
\label{sec:set_gen}
CNN-based object detectors require large amounts of annotated data for training, due to the large number of parameters that need to be learned. For object instance detection the training data should also cover the variations in the object's viewpoint and other nuisance parameters such as lighting, occlusion and clutter.
Manually collecting and annotating scenes with the aforementioned properties is time-consuming  and costly. 
Another factor in annotation is the sometimes low generalization capability of trained models across different environments and backgrounds. The work of~\cite{Xiao_Robot} addressed this problem by building a map of an environment including objects of interest and using Amazon Mechanical Turk for annotation and subsequent training of object detectors in each particular environment. 
The authors demonstrated this approach on commonly encountered categories ($\approx$ 20) of household objects. This approach uses human labeling effort for each new each scene and object combination, potentially limiting scalability. 

Our approach focuses on object instances and their superimposition into real scenes at different positions, scales, while reducing the difference in lighting conditions and exploiting proper context. To this end, we use cropped images from existing object recognition datasets such as BigBird~\cite{Singh_ICRA14} rather than using 3D CAD models~\cite{SaenkoICCV15,Su_ICCV15}. This allows us to have real colors and textures for our training instances as opposed to rendering them with randomly chosen or artificial samples. The BigBird dataset captures 120 azimuth angles from 5 different elevations for a total of 600 views per object. It contains a total of 125 object instances with a variety of textures and shapes. In our experiments we use the 11 object instances that can be found in the GMU-Kitchens dataset.



The process of generating a composite image with superimposed objects can be summarized in the following steps. First, we choose a background scene and estimate the positions of any support surfaces. This is further augmented by semantic segmentation of the scene, used to verify the support surfaces found by plane fitting.  The objects of interest are placed on support surfaces, ensuring their location in areas with appropriate context and backgrounds. The next step is to randomly choose an object and its pose, followed by choosing a position in the image. The object scale is then determined by the depth value of the chosen position and finally the object is blended into the scene. An example of this process is shown in Figure~\ref{fig:blending_procedure}. We next describe these steps in more detail. 

\paragraph*{Selective Positioning} In natural images, small hand-held objects are usually found on supporting surfaces such as counters, tables, and desks. 
These planar surfaces are extracted using the method described in~\cite{Taylor_RSS12}, which applies RANSAC to fit planes to regions after an initial over-segmentation of the image. Given the extracted planar surfaces's orientations, we select the planes with large extent, which are aligned with the gravity direction as candidate support surfaces. To ensure that the candidate support surfaces belong to a desired semantic category, a support surface is considered valid if it overlaps in the image with semantic categories of counters, tables and desks obtained by semantic segmentation of the RGB-D image.

\paragraph*{Semantic Segmentation} To determine the semantic categories in the scene, we use the semantic segmentation CNN of~\cite{Arsalan_3DV16}, which is pre-trained on MS-COCO and PASCAL-VOC datasets, and fine-tuned on NYU Depth v2 dataset for 40 semantic categories. The model is jointly trained for semantic segmentation and depth estimation, which allows the scene geometry to be exploited for better discrimination between some of the categories. We do not rely solely on the semantic segmentation for object positioning, since it rarely covers the entire support surface, as can be seen in Figure~\ref{fig:blending_procedure}(c). The combination of the support surface detection and semantic segmentation produces more accurate regions for placing the objects. The aforementioned regions that belong to valid support surfaces are then randomly chosen for object positioning. Finally, occlusion levels are regulated by allowing a maximum of 40\% overlap between positioned object instances in the image.



\paragraph*{Selective Scaling and Blending}
The size of the object is determined by using the depth of the selected position and scaling the width $w$ and height $h$ accordingly:
\begin{align*}
\hat{w}&=\frac{w\bar{z}}{z} & \hat{h}&=\frac{h\bar{z}}{z}
\end{align*}
where $\bar{z}$ is the median depth of the object's training images, $z$ is the depth at the selected position in the background image, and $\hat{w}$, $\hat{h}$ are the scaled width and height respectively. 

\begin{figure*}[t] 
      \centering
      \includegraphics[width=0.2\textwidth]{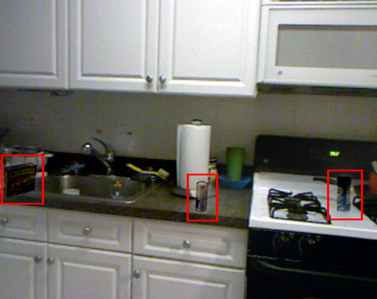}
      \includegraphics[width=0.2\textwidth]{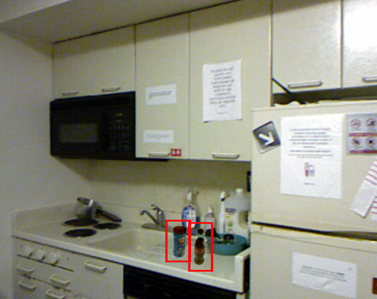}
      \includegraphics[width=0.2\textwidth]{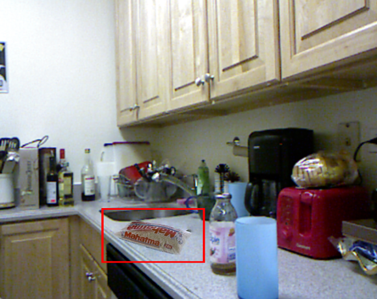}
      \includegraphics[width=0.2\textwidth]{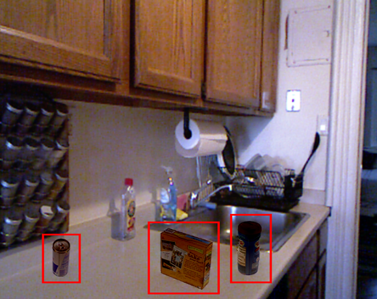}
      \\
      \vspace{1mm}
	  \includegraphics[width=0.2\textwidth]{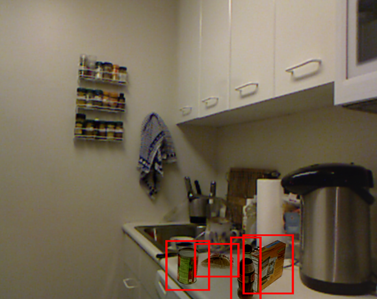}
	  \includegraphics[width=0.2\textwidth]{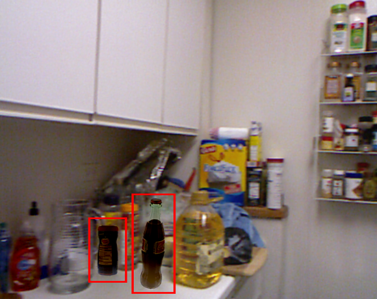}
      \includegraphics[width=0.2\textwidth]{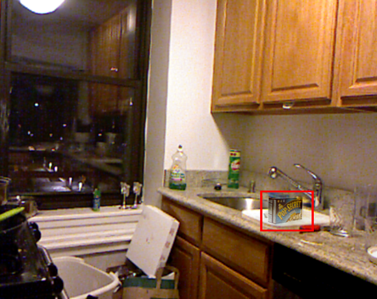}
      \includegraphics[width=0.2\textwidth]{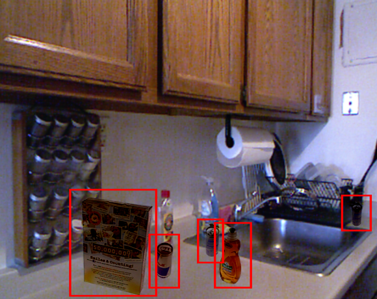}
      \\
      \vspace{1mm}
	  \includegraphics[width=0.2\textwidth]{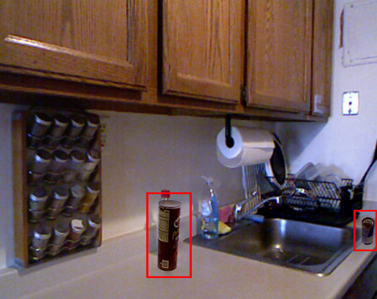}
	  \includegraphics[width=0.2\textwidth]{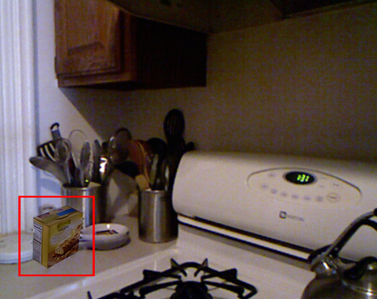}
      \includegraphics[width=0.2\textwidth]{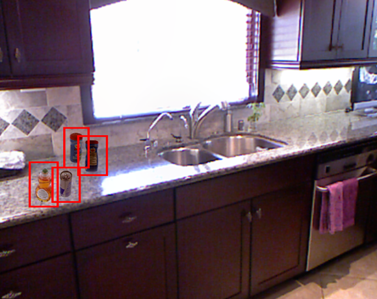}
      \includegraphics[width=0.2\textwidth]{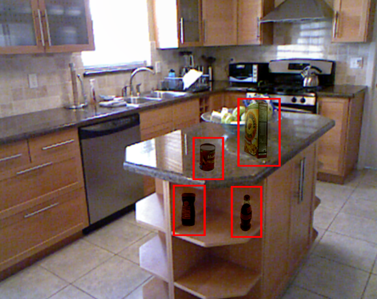}
      \caption{Examples of blending object instances from the BigBird dataset into scenes from the NYU Depth V2 dataset. The blended objects are marked with a red bounding box. Best viewed in color.}
      \label{fig:blending_examples}
\end{figure*}

The last step in our process is to blend the object with the background image in order to mitigate the effects of changes in illumination and contrast. We use the implementation from Fast Seamless Cloning~\cite{Tanaka_SIGGRAPH12} with a minor modification. Instead of blending a rectangular patch of the object, we provide a masked object to the fast seamless cloning algorithm which produces a cleaner result. Figure~\ref{fig:blending_examples} illustrates examples of scenes with multiple blended objects.

\subsection{Object Detectors}
For our experiments we employ two state-of-the-art object detectors, Faster R-CNN~\cite{Ren_NIPS15} and Single-Shot Multibox Detector (SSD)~\cite{Liu_ECCV16}. Both Faster R-CNN and SSD are trained end-to-end but their architectures are different. Faster R-CNN consists of two modules. The first module is the Region Proposal Network (RPN) which is a fully convolutional network that outputs object proposals and also an objectness score for each proposal reflecting the probability of having an object inside the region. The second detection network module resizes the feature maps, corresponding to each object proposal to a fixed size, classifies it to an object category and refines the location and the height and width of the bounding box associated with each proposal. 
The advantage of Faster R-CNN is the modularity of the model; one module that finds object proposals and the second module which classifies each of the proposals. The downside of Faster R-CNN is that it uses the same feature map to find objects of different sizes which causes problems for small objects. SSD tackles this problem by creating feature maps of different resolutions. Each cell of the coarser feature maps captures larger area of the image for detecting large objects whereas the finer feature maps are detecting smaller objects. These multiple feature maps allow higher accuracy for a given input resolution, providing SSD's speed advantage for similar accuracy. Both detectors have difficulties for objects with small size in pixels, making input resolution an important factor. 

\begin{figure}[t] 
      \centering
      \includegraphics[width=0.15\textwidth,height=0.17\textwidth]{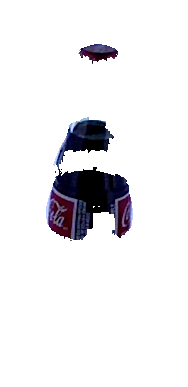}
      \includegraphics[width=0.15\textwidth,height=0.17\textwidth]{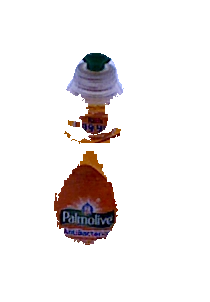}
      \includegraphics[width=0.15\textwidth,height=0.17\textwidth]{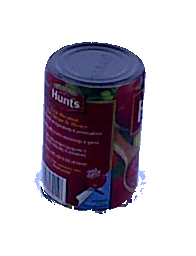}
      \\
      \vspace{1mm}
      \includegraphics[width=0.15\textwidth,height=0.17\textwidth]{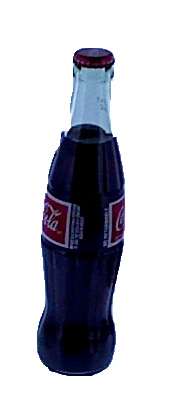}
      \includegraphics[width=0.15\textwidth,height=0.17\textwidth]{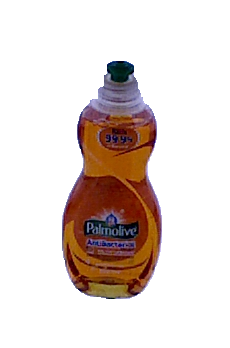}
      \includegraphics[width=0.15\textwidth,height=0.17\textwidth]{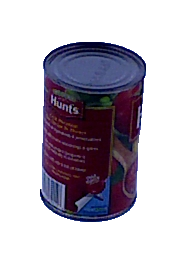}
      \caption{Comparison between masks from BigBird (top row), and masks after refinement with Graph-cut (bottom row).}
      \label{fig:mask_comparison}
\end{figure}


\begin{table*}[t]
\begin{center}
\resizebox{18cm}{!}{
	\begin{tabular}{l|c|c|c|c|c|c|c|c|c|c|c||c}
    & \rot{coca cola} & \rot{coffee mate} & \rot{honey bunches} & \rot{hunts sauce} & \rot{mahatma rice} & \rot{nature valley 1} & \rot{nature valley 2} & \rot{palmolive orange} & \rot{pop secret} & \rot{pringles bbq} & \rot{red bull} & \rot{mAP} \\
    \hline \hline
    \multicolumn{13}{c}{\textbf{Real to Real}} \\
    \hline
 1. \textbf{3-Fold} & 46.7 / 78.9 & 73.7 / 92.0 & 81.8 / 91.9 & 64.5 / 81.8 & 62.9 / 74.7 & 83.9 / 93.4 & 70.3 / 85.9 & 69.8 / 76.6 & 76.1 / 90.7 & 64.8 / 86.4 & 27.7 / 54.6 & 65.6 / 82.5 \\
    \hline \hline
	\multicolumn{13}{c}{\textbf{Synthetic to Real}} \\
	\hline
 2. \textbf{RP-SI-RS} & 9.1 / 37.2 & 15.9 / 68.3 & 49.4 / \textbf{72.5} & 9.8 / 26.1 & \textbf{13.5} / 32.3 & 61.6 / 70.2 & 42.2 / 57.2 & 15.5 / 29.0 & 36.9 / 46.9 & 1.0 / 2.9 & 0.8 / 20.4 & 23.2 / 42.1 \\
    \hline
 3. \textbf{RP-BL-RS} & 9.1 / \textbf{62.4} & 15.7 / 69.3 & 41.4 / 58.2 & \textbf{10.7} / 32.4 & 9.9 / 4.3 & 41.2 / 51.7 & 39.0 / 47.1 & \textbf{28.7} / \textbf{39.3} & 36.2 / 32.2 & 11.5 / \textbf{59.2} & 0.7 / \textbf{30.0} & 22.2 / 44.2 \\
    \hline
 4. \textbf{SP-SI-SS} & 10.5 / 45.2 & 17.5 / \textbf{71.7} & 47.2 / 66.6 & 0.1 / 26.0 & 9.1 / \textbf{45.5} & 44.9 / 80.5 & 36.5 / \textbf{78.4} & 24.0 / 37.8 & 9.1 / 46.1 & 5.5 / 27.1 & \textbf{10.3} / 9.7 & 19.5 / 48.6 \\
    \hline
 5. \textbf{SP-BL-SS} & \textbf{18.3} / 55.5 & \textbf{22.1} / 67.9 & \textbf{58.9} / 71.2 & 9.5 / \textbf{34.6} & 11.1 / 30.6 & \textbf{75.7} / \textbf{82.9} & \textbf{65.5} / 66.2 & 23.8 / 33.1 & \textbf{59.4} / \textbf{54.3} & \textbf{14.6} / 54.8 & 9.1 / 17.7 & \textbf{33.5} / \textbf{51.7} \\
    \hline \hline
    \multicolumn{13}{c}{\textbf{Synthetic+Real to Real}} \\
    \hline
 6. \textbf{1\% real} & 39.4 / 65.1 & 71.8 / 85.8 & 80.4 / 85.7 & 50.2 / 62.3 & 45.2 / 51.6 & 82.6 / 90.4 & 74.9 / 85.6 & 57.8 / 54.3 & 78.1 / 79.4 & 54.2 / 70.6 & 28.5 / 32.2 & 60.3 / 69.3 \\
    \hline
  7. \textbf{10\% real} & 59.4 / 70.5 & 83.8 / 91.5 & 83.7 / 89.6 & 66.2 / 82.2 & 60.7 / 62.8 & 87.3 / 94.6 & \textbf{79.8} / 87.4 & 72.6 / 66.3 & 83.4 / 89.5 & 77.6 / 87.4 & 33.0 / 49.5 & 71.6 / 79.2 \\
    \hline
 8. \textbf{50\% real} & \textbf{64.6} / 79.3 & 84.2 / 92.5 & \textbf{87.6} / 91.1 & 70.4 / 77.3 & 67.1 / \textbf{86.2} & \textbf{89.2} / 95.4 & 79.7 / 87.9 & \textbf{75.4} / 77.8 & 80.1 / 91.6 & 79.3 / 90.1 & \textbf{37.6} / 52.2 & \textbf{74.1} / 83.8 \\
    \hline
   9. \textbf{100\% real} & 59.0 / \textbf{82.6} & \textbf{84.5} / \textbf{92.9} & 85.1 / \textbf{91.4} & \textbf{74.2} / \textbf{85.5} & \textbf{67.5} / 81.9 & 87.4 / \textbf{95.5} & 78.9 / \textbf{88.6} & 71.3 / \textbf{78.5} & \textbf{85.2} / \textbf{93.6} & \textbf{79.9} / \textbf{90.2} & \textbf{37.6} / \textbf{54.1} & 73.7 / \textbf{85.0} \\
    \hline \hline
    \multicolumn{13}{c}{\textbf{Synthetic to Synthetic}} \\
    \hline
 10. \textbf{RP-SI-RS} & \textbf{90.8} / \textbf{99.6} & \textbf{90.9} / \textbf{100} & \textbf{90.8} / \textbf{99.7} & \textbf{90.8} / \textbf{99.6} & \textbf{90.9} / \textbf{99.6} & \textbf{90.9} / \textbf{99.8} & \textbf{90.8} / \textbf{99.7} & \textbf{90.7} / \textbf{98.9} & \textbf{90.9} / \textbf{99.7} & \textbf{90.8} / \textbf{99.4} & \textbf{90.6} / \textbf{98.7} & \textbf{90.8} / \textbf{99.5}\\
    \hline
 11. \textbf{SP-BL-SS} & 84.3 / 79.2 & 86.7 / 84.4 & 88.1 / 94.8 & 81.7 / 79.3 & 88.9 / 94.6 & 83.5 / 92.6 & 80.8 / 89.5 & 83.1 / 79.9 & 84.5 / 93.1 & 86.4 / 89.1 & 74.0 / 65.8 & 83.8 / 85.7 \\
    \hline  \hline
	\end{tabular}
	}
    \caption{Average precision results (SSD / Faster R-CNN) for all experiments on the GMU-Kitchens dataset. The Synthetic+Real to Real experiments were performed using the SP-BL-SS set plus the percentage of real data shown in the table.}
    \label{tab:results_gmuk}
\end{center}
\end{table*}

\begin{table*}[t]
\begin{center}
	\begin{tabular}{l|c|c|c|c|c||c}
	 & \rot{bowl} & \rot{cap} & \rot{cereal box} & \rot{coffee mug} & \rot{soda can} & \rot{mAP} \\
    \hline \hline
    \multicolumn{7}{c}{\textbf{Real to Real}} \\
    \hline
    1.\textbf{Scenes 1-7} & 90.8 / 99.7 & 90.2 / 95.5 & 90.9 / 99.6 & 89.9 / 96.7 & 89.3 / 96.7 & 90.2 / 97.9\\
    \hline \hline
    \multicolumn{7}{c}{\textbf{Synthetic to Real}} \\
    \hline
   2. \textbf{RP-SI-RS} & 77.2 / 65.2 & 78.5 / 39.4 & \textbf{90.9} / 69.2 & 74.1 / 57.0 & 70.1 / 29.4 & 78.2 / 52.0 \\
    \hline
    3. \textbf{RP-BL-RS} & 71.5 / 82.6 & 62.9 / 47.3 & 90.3 / 93.0 & 73.1 / \textbf{74.9} & 65.2 / 52.7 & 72.6 / 70.1 \\
    \hline
   4. \textbf{SP-SI-SS} & \textbf{77.8} / \textbf{86.6} & \textbf{79.8} / 62.4 & 90.8 / 93.6 & 73.4 / 68.6 & \textbf{75.5} / 55.5 & \textbf{79.5} / 73.4 \\
    \hline
   5. \textbf{SP-BL-SS} & 71.9 / 82.3 & 75.9 / \textbf{70.9} & 90.7 / \textbf{96.8} & \textbf{74.3} / 74.2 & 75.0 / \textbf{66.3} & 77.5 / \textbf{78.1}\\
    \hline \hline
    \multicolumn{7}{c}{\textbf{Synthetic+Real to Real}} \\
    \hline
   6. \textbf{1\% real} & 87.7 / 98.3 & 88.3 / 92.7 & 90.8 / 98.6 & 88.1 / 96.6 & 89.5 / 94.9 & 88.9 / 96.2\\
    \hline
  7. \textbf{10\% real} & 90.8 / \textbf{99.6} & 89.5 / 96.0 & 90.8 / 99.6 & 90.4 / 96.9 & \textbf{90.8} / 97.1 & 90.5 / 97.8\\
    \hline
  8. \textbf{50\% real} & \textbf{90.9} / 99.5 & \textbf{90.6} / 96.5 & \textbf{90.9} / \textbf{99.9} & 90.3 / \textbf{97.3} & 90.6 / 97.8 & 90.7 / \textbf{98.2}\\
    \hline
   9. \textbf{100\% real} & \textbf{90.9} / 99.4 & 90.5 / \textbf{97.0} & \textbf{90.9} / 99.3 & \textbf{90.8} / 97.2 & \textbf{90.8} / \textbf{98.1} & \textbf{90.8} / \textbf{98.2}\\
    \hline \hline    
    \multicolumn{7}{c}{\textbf{Synthetic to Synthetic}} \\
    \hline
  10. \textbf{RP-SI-RS} & 90.5 / \textbf{98.5} & \textbf{90.9} / \textbf{99.5} & \textbf{90.9} / \textbf{99.5} & \textbf{90.2} / \textbf{96.9} & \textbf{90.0} / 92.6 & \textbf{90.5} / \textbf{97.4} \\
    \hline
  11. \textbf{SP-BL-SS} & \textbf{90.7} / 97.4 & 90.7 / 97.5 & 90.4 / 97.3 & 89.5 / 95.1 & 89.2 / \textbf{93.5} & 90.1 / 96.2 \\
    \hline \hline
  	\end{tabular}
    \caption{Average precision results (SSD / Faster R-CNN) for all experiments on the WRGB-D dataset. The Synthetic+Real to Real experiments were performed using the SP-BL-SS set plus the percentage of real data shown in the table.}
    \label{tab:results_wrgbd}
\end{center}
\end{table*}

\begin{figure*}[t] 
      \centering
      \includegraphics[width=0.92\textwidth]{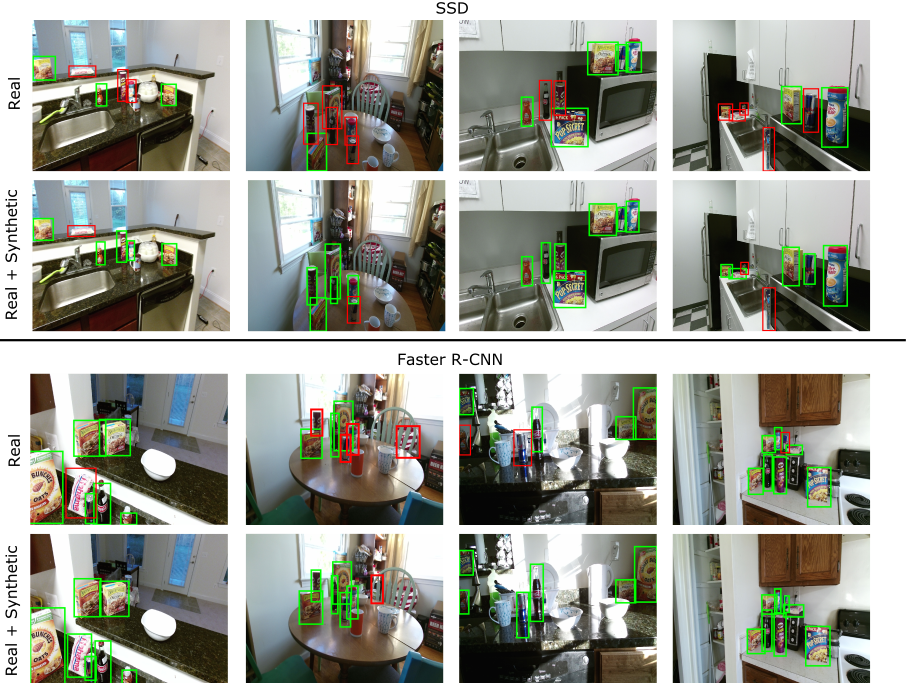}
      \caption{Detection examples for the SSD and Faster R-CNN object detectors on the GMU-Kitchens dataset. Rows 1 and 3 show results when only the real training data were used, while rows 2 and 4 present results after the detectors were trained with the synthetic set \textbf{SP-BL-SS} and 50\% of the real training data. The green bounding boxes depict correct detections, while the red represent false classifications and missed detections. Training with a combination of synthetic and real data proves beneficial for the detection task, as the detectors are more robust to small objects and viewpoint variation. Best viewed in color.}
      \label{fig:detection_examples}
\end{figure*}

\section{Experiments}

In order to evaluate the object detectors trained on composited images, we have conducted three sets of experiments on two publicly available datasets, the GMU-Kitchen Scenes ~\cite{Georgakis_3DV16} and the Washington RGB-D Scenes v2 dataset~\cite{Lai_icra14}. In the first experiment, training images are generated by choosing different compositing strategies to determine the effect of positioning, scaling, and blending on the performance. The object detectors are trained on composited images and evaluated on real scenes. In the second set of experiments we examine the effect of varying proportion of synthetic/composited images and real training images. Finally we use synthetic data for both training and testing in order to show the reduction of over-fitting to superimposition artifacts during training when the proposed approach of data generation is employed.

The code for the synthesization process along with the background scenes and synthetic data are available at: \url{http://cs.gmu.edu/~robot/synthesizing.html}.

\subsection{Datasets and Backgrounds}
For our experiments, we utilized the following datasets:
\paragraph{GMU Kitchen Scenes dataset ~\cite{Georgakis_3DV16}}
The GMU-Kitchens dataset includes 9 RGB-D videos of kitchen scenes with 11 object instances from the BigBird dataset. We also used all 71 raw kitchen videos from the NYU Depth Dataset V2~\cite{Silberman_ECCV12} with a total of around 7000 frames as background images. For each image we generate four synthetic images with different variations in objects that are added to the scene, pose, scale, and the location that the objects are put. The object identities and their poses are randomly sampled from the BigBird dataset, from 360 examples per object with 3 elevations and 120 azimuths. 


The images where the support surfaces were not detected are removed from the training set, making our effective set around 5000 background images. Cropped object images from BigBird dataset of the 11 instances contained in GMU-Kitchens were used for superimpositioning. We refine the provided object masks with GraphCut~\cite{Boykov_TPAMI01}, in order to get cleaner outlines for the objects. This helps with the jagged and incomplete boundaries of certain objects (e.g. coke bottle), which are due to imperfect masks obtained from the depth channel of RGB-D data caused by reflective materials. Figure~\ref{fig:mask_comparison} illustrates a comparison between masks from BigBird and masks refined with GraphCut algorithm.
For comparison with the rest of the experiments we also provide the performance of the object detectors (row 1 of Table~\ref{tab:results_gmuk}) trained and tested on the real data. The train-test split follows the division of the dataset into three different folds. In each fold six scenes are used for training and three are used for testing, as shown in~\cite{Georgakis_3DV16}. 

\paragraph{Washington RGB-D Scenes v2 dataset (WRGB-D)~\cite{Lai_icra14}}
The WRGB-D dataset includes 14 RGB-D videos of indoor table-top scenes containing instances of objects from five object categories: bowl, cap, cereal box, coffee mug, and soda can. The synthetic training data is generated using the provided background scenes (around 3000 images) and cropped object images for the present object categories in the WRGB-D v1 dataset~\cite{Lai_ICRA11}. 
For each background image we generate five synthetic images to get a total of around 4600 images. As mentioned earlier, images without a support surface are discarded.  The images that belong to seven of these scenes are used for training and the rest is used for testing. Line 1 in Table~\ref{tab:results_wrgbd} shows the performance of the two object detectors with this split of the real training data.



\subsection{Synthetic to Real}
\label{sec:synthetic_to_real}
In this experiment we use the synthetic training sets generated with different combinations of generation parameters for training, and test on real data. The generation parameters that we vary are: Random Positioning (\textbf{RP}) / Selective Positioning (\textbf{SP}), Simple Superimposition (\textbf{SI}) / Blending (\textbf{BL}), and Random Scale (\textbf{RS}) / Selective Scale (\textbf{SS}), where \textbf{SP}, \textbf{SS}, and \textbf{BL} are explained in Section~\ref{sec:set_gen}. For \textbf{RP} we randomly sample the position for the object in the entire image, for \textbf{RS} the scale of the object is randomly sampled from the range of 0.2 to 1 with a step of 0.1, and for \textbf{SI} we do not use blending but instead we superimpose the masked object directly on the background. 

The objective of this experiment is to investigate the effect of the generation parameters on the detection accuracy. For example, if a detector is trained on a set generated with selective positioning, with blending, and selective scale, how does it compare to another detector which is trained on a completely randomly generated set with blending? If the former demonstrates higher performance than the latter, then we can assume that selective positioning and scaling are important and superior to random positioning. For each trained detector, a combination of the generation parameters (e.g. \textbf{SP-BL-SS}) is chosen, and then the synthetic set is generated using our proposed approach along with its bounding box annotations for each object instance. The detector is trained only on the synthetic data and then tested on the real data. 

The results are shown on lines 2-5 in Table~\ref{tab:results_gmuk} for the GMU-Kitchens dataset and in Table~\ref{tab:results_wrgbd} for the WRGB-D dataset. Note that for the GMU-Kitchens dataset, all frames from 9 scenes videos were used for testing. We report detection accuracy on four combinations of generation parameters, \textbf{RP-SI-RS}, \textbf{RP-BL-RS}, \textbf{SP-SI-SS}, and \textbf{SP-BL-SS}. Other combinations such as \textbf{SP-BL-RS} and \textbf{RP-BL-SS} have also been tried, however we noticed that applying selective positioning without selective scaling and vice-versa, does not yield any significant improvements.

For both datasets, we first notice that using only synthetic data for training considerably lowers the detection accuracy compared to using real training data.
Nevertheless, when training with synthetic data, the \textbf{SP-BL-SS} generation approach produced an improvement of 10.3\% and 9.6\% for SSD and Faster R-CNN respectively over the randomized generation approach, \textbf{RP-SI-RS}, on the GMU-Kitchens dataset. This suggests that selective positioning and scaling are important factors when generating the training set. 

In the case of the WRGB-D dataset, different blending strategies work better for SSD and Faster R-CNN, \textbf{SP-SI-SS} and \textbf{SP-BL-SS} respectively. The right choice of blending strategy seems to improve Faster R-CNN somewhat more, while the overall performance of the two detectors is comparable. The positioning strategy, {\bf SP} vs {\bf RP}, affects the two detectors differently on this dataset.  
SSD achieves higher performance with the random positioning \textbf{RP-SI-RS}, while Faster R-CNN shows a large improvement of 26.1\% when it is trained with \textbf{SP-BL-SS}. This can be explained by the fact that Faster R-CNN is trained on proposals from the Region Proposal Network (RPN), which under-performs when objects are placed randomly in the image (as in \textbf{RP-SI-RS}). On the other hand, SSD does not have any prior knowledge about the location of the objects so it learns to regress the bounding boxes from scratch. The bounding boxes in the beginning of the training are generated randomly until the SSD learns to localize the objects. This trend is not observed for the GMU-Kitchens dataset since it has more clutter in the scenes and higher variability of backgrounds, which makes the localization of the objects harder. To justify this argument, we performed a side-experiment where we run the pre-trained RPN on both WRGB-D and GMU-Kitchens dataset and evaluated in terms or recall. Results can be seen in Table~\ref{tab:rpn}, where RPN performs much better on the WRGB-D dataset than on GMU-Kitchens.

\begin{table}[t]
\begin{center}
	\begin{tabular}{c|c|c}
    \hline
	IoU & GMU-Kitchens & WRGB-D \\
    \hline
    0.5 & 76.6 & 98.0 \\
    \hline
    0.7 & 28.6 & 60.8 \\
    \hline
  	\end{tabular}
    \caption{Recall (\%) results for the RPN on the GMU-Kitchens and WRGB-D datasets on two different Intersection over Union (IoU) thresholds. In all cases, RPN generated 3000 proposals per image.}
    \label{tab:rpn}
\end{center}
\end{table}




\subsection{Synthetic+Real to Real}
We are interested to see how effective our synthetic training set is when combined with real training data. Towards this end the two detectors are trained using the synthetic set with selective positions and blending \textbf{SP-BL-SS} with certain percentage of the real training data: 1\%, 10\%, 50\%, and 100\%. For the real training data, besides the case of 100\%, the images are chosen randomly. 

Results are shown in lines 6-9 in Table~\ref{tab:results_gmuk} for the GMU-Kitchens dataset and in Table~\ref{tab:results_wrgbd} for the WRGB-D dataset. {\em What is surprising in these results} is that when synthetic training data is combined with only 10\% of the real training data, we achieve higher or comparable detection accuracy than when the training set is only comprised with real data (see line 1 in both tables). In the case of SSD in the GMU-Kitchens dataset, we observe an increase of 6\%. Only exception is Faster R-CNN on the GMU-Kitchens dataset which achieves a 2.3\% lower performance, however, when we use 50\% of the real training data we get a better performance of 1.3\%. In all cases, when the synthetic set is combined with 50\% and 100\% of the real data, it outperforms the training with the real training set.

The results suggest that our synthetic training data can effectively augment existing datasets even when the actual number of real training examples is small. This is particularly useful when only a small subset of the data is annotated. Specifically, in our settings, the 10\% of real training data refers to around 400 images in the GMU-Kitchens dataset, and  around 600 in the WRGB-D dataset. Figure~\ref{fig:detection_examples} presents examples for which the detectors were unable to detect objects when they were trained with only real data, but succeeded when the training set was augmented with our synthetic data.

We further support our argument by comparing the performance of the detectors trained only on varying percentages of the real data to being trained by real+synthetic in Table~\ref{tab:real_per}. The synthetic set here is also generated using \textbf{SP-BL-SS}. Note that for most of the cases the accuracy increases when the detectors are trained with both real and synthetic data, and the largest gain is observed for SSD.

Finally, we present results for the GMU-Kitchens dataset when the percentage of synthetic data (\textbf{SP-BL-SS}) is varied, while the real training data remains constant, in Table~\ref{tab:synth_per}. Again, SSD shows a large and continuing improvement as the amount of the synthetic data increases, while Faster R-CNN achieves top performance when half of the synthetic data are used for training.

\begin{table}[t]
\begin{center}
	\begin{tabular}{c|c|c}
    \multicolumn{3}{c}{\textbf{GMU-Kitchens}} \\
    \hline
	 Percentage of Real & Only Real & Real+Synthetic \\
    \hline
    1\% & 57.4 / \textbf{70.8} & \textbf{60.3} / 69.3 \\
    \hline
    10\% & 61.5 / \textbf{81.1} & \textbf{71.6} / 79.2 \\
    \hline
    50\% & 66.4 / 82.4 & \textbf{74.1} / \textbf{83.8} \\
    \hline
    100\% & 65.6 / 82.5 & \textbf{73.7} / \textbf{85.0} \\
    \hline \hline
	\multicolumn{3}{c}{\textbf{WRGB-D}} \\
    \hline
	Percentage of Real & Only Real & Real+Synthetic \\
    \hline
    1\% & \textbf{89.1} / 95.6 & 88.9 / \textbf{96.2} \\
    \hline
    10\% & 89.4 / 97.5 & \textbf{90.5} / \textbf{97.8} \\
    \hline
    50\% & 90.2 / 97.6 & \textbf{90.7} / \textbf{98.2} \\
    \hline
    100\% & 90.2 / 97.9 & \textbf{90.8} / \textbf{98.2} \\
    \hline
  	\end{tabular}
    \caption{Comparison in performance of SSD / Faster R-CNN between training with only real to training with real+synthetic, with varying amounts of real data. The amount of synthetic data is constant.}
    \label{tab:real_per}
\end{center}
\end{table}

\begin{table}[t]
\begin{center}
	\begin{tabular}{c|c}
    \hline
	Training Set & Accuracy \\
    \hline
    Real & 65.6 / 82.5 \\
    \hline
    Real+Synth(10\%) & 69.0 / 84.5 \\
    \hline
    Real+Synth(50\%) & 72.7 / \textbf{85.7} \\
    \hline
    Real+Synth(100\%) & \textbf{73.7} / 85.0 \\
    \hline
  	\end{tabular}
    \caption{Comparison in performance of SSD / Faster R-CNN on the GMU-Kitchens dataset for increasing amounts of the synthetic data, while all real data are used.}
    \label{tab:synth_per}
\end{center}
\end{table}

\subsection{Synthetic to Synthetic}
In this experiment, the object detectors are trained and tested on synthetic sets. The objective is to show the reduction of over-fitting on the training data when using our approach to generate the synthetic images, instead of creating them randomly. We used the synthetic sets of \textbf{RP-SI-RS} and \textbf{SP-BL-SS} and split them in half in order to create the train-test sets.

The results are presented on lines 10 and 11 in Table~\ref{tab:results_gmuk} for the GMU-Kitchens dataset and in Table~\ref{tab:results_wrgbd} for the WRGB-D dataset. For GMU-Kitchens, we observe that \textbf{RP-SI-RS} achieves results of over 90\%, and in the case of Faster R-CNN almost 100\%, while at the same time it is the least performing synthetic set in the synthetic to real experiment (see line 2 in table~\ref{tab:results_gmuk}) described in Section~\ref{sec:synthetic_to_real}. This is because the detectors over-fit on the synthetic data and cannot generalize to an unseen set of real test data. While the detectors still seem to over-fit on \textbf{SP-BL-SS}, the gap between the accuracy on the synthetic testing and real testing data is much smaller, at the order of 17.3\% for SSD, and 23.4\% for Faster R-CNN (see line 5 in table~\ref{tab:results_gmuk}). 

On the other hand, for the WRGB-D dataset both synthetic training sets achieve similar results on their synthetic test sets. This is not surprising as the complexity of the scenes is much lower in WRGB-D than in the GMU-Kitchens dataset. Please see section~\ref{sec:synthetic_to_real} for more details.  


\subsection{Additional Discussion}

We have seen in the results of section~\ref{sec:synthetic_to_real}, that when a detector is trained on synthetic data and then applied on real data, the performance is consistently lower that training on real data. While this can be attributed to artifacts introduced during the blending process, one other factor is the large difference of backgrounds between the NYU V2 dataset and the GMU-Kitchens. We investigated this through a simple object recognition experiment. We trained the VGG~\cite{Simonyan_arXiv14} network on the BigBird dataset on the cropped images with elevation angles from cameras 1, 3, and 5, tested on the images with elevation angles from cameras 2 and 4, and achieved recognition accuracy of 98.2\%. For comparison, when the VGG is trained on all images from BigBird, and tested on cropped images from the GMU-Kitchens, which contain real background scenes, the accuracy drops down to 79.0\%.

\section{Conclusion}
One of the advantages of our method is that it is scalable both with the number of objects of interest and with the set of the possible backgrounds, which makes our method suitable for robotics application. For example, the object detectors can be trained with significantly less annotated data using our proposed training data augmentation. We also showed that our method is more effective when the object placements are based on semantic and geometric context of the scene. This is due to the fact that CNNs implicitly consider the surrounding context of the objects and when superimposition is informed by semantic and geometric factors, the gain in accuracy is larger.
Another related observation is that for SSD, accuracy  increases more than for Faster R-CNN when training data is augmented by synthetic composite images. 

While we showed it is possible to train an object detector with fewer annotated images using synthetically generated images, alternative domain adaptation approaches can be also explored towards the goal of reducing the amount of human annotation required. 


In conclusion, we have presented an automated procedure for generating synthetic training data for deep CNN object detectors. The generation procedure takes into consideration geometry and semantic segmentation of the scene in order to make informed decisions regarding the positions and scales of the objects. We have employed two state-of-the-art object detectors and demonstrated an increase in their performance when they are trained with an augmented training set. In addition, we also investigated the effect of different generation parameters and provided some insight that could prove useful in future attempts to generate synthetic data for training object detectors.

\section*{Acknowledgments}
We acknowledge support from NSF NRI grant 1527208. Some of the experiments were run on ARGO, a research computing cluster provided by the Office of Research Computing at George Mason University, VA. (URL: http://orc.gmu.edu).


\bibliographystyle{plainnat}
\bibliography{references}

\begin{thebibliography}{28}
\providecommand{\natexlab}[1]{#1}
\providecommand{\url}[1]{\texttt{#1}}
\expandafter\ifx\csname urlstyle\endcsname\relax
  \providecommand{\doi}[1]{doi: #1}\else
  \providecommand{\doi}{doi: \begingroup \urlstyle{rm}\Url}\fi

\bibitem[Boykov et~al.(2001)Boykov, Veksler, and Zabih]{Boykov_TPAMI01}
Y.~Boykov, O.~Veksler, and R.~Zabih.
\newblock Fast approximate energy minimization via graph cuts.
\newblock \emph{IEEE Transactions on Pattern Analysis and Machine Intelligence
  (PAMI)}, 23\penalty0 (11):\penalty0 1222--1239, November 2001.
\newblock ISSN 0162-8828.
\newblock \doi{10.1109/34.969114}.
\newblock URL \url{http://dx.doi.org/10.1109/34.969114}.

\bibitem[Cheng et~al.(2014)Cheng, Zhang, Lin, and Torr]{Cheng_CVPR14}
M.~M. Cheng, Z.~Zhang, W.~Y. Lin, and P.~Torr.
\newblock {BING}: Binarized normed gradients for objectness estimation at
  300fps.
\newblock In \emph{IEEE Conference on Computer Vision and Pattern Recognition
  (CVPR)}, 2014.

\bibitem[Collet et~al.(2011)Collet, Martinez, and Srinivasa]{collet_IJRR11}
A.~Collet, M.~Martinez, and S.~Srinivasa.
\newblock The {MOPED} framework: Object recognition and pose estimation for
  manipulation.
\newblock In \emph{International Journal of Robotics Research (IJRR)}, 2011.

\bibitem[Felzenszwalb et~al.(2010)Felzenszwalb, Girshick, McAllester, and
  Ramanan]{Felzenszwalb_TPAMI10}
P.~Felzenszwalb, R.~Girshick, D.~McAllester, and D.~Ramanan.
\newblock Object detection with discriminatively trained part based models.
\newblock \emph{IEEE Transactions on Pattern Analysis and Machine Intelligence
  (PAMI)}, 32\penalty0 (9):\penalty0 1627--1645, 2010.

\bibitem[Georgakis et~al.(2016)Georgakis, Reza, Mousavian, Le, and
  Kosecka]{Georgakis_3DV16}
G.~Georgakis, Md. Reza, A.~Mousavian, P.~Le, and J.~Kosecka.
\newblock Multiview {RGB-D} dataset for object instance detection.
\newblock In \emph{IEEE International Conference on 3DVision (3DV)}, 2016.

\bibitem[Girshick(2015)]{Girshick_ICCV15}
R.~Girshick.
\newblock Fast {R-CNN}.
\newblock In \emph{{IEEE} International Conference on Computer Vision (ICCV)},
  2015.

\bibitem[Girshick et~al.(2014)Girshick, Donahue, Darrell, and
  Malik]{Girshick_CVPR14}
R.~Girshick, J.~Donahue, T.~Darrell, and J.~Malik.
\newblock Rich feature hierarchies for accurate object detection and semantic
  segmentation.
\newblock In \emph{{IEEE} Conference Computer Vision and Pattern Recognition
  (CVPR)}, 2014.

\bibitem[Gupta et~al.(2016)Gupta, Vedaldi, and Zisserman]{Gupta_CVPR16}
A.~Gupta, A.~Vedaldi, and A.~Zisserman.
\newblock Synthetic data for text localisation in natural images.
\newblock In \emph{IEEE Conference on Computer Vision and Pattern Recognition
  (CVPR)}, 2016.

\bibitem[Krizhevsky et~al.(2012)Krizhevsky, Sutskever, and
  Hinton]{Krizhevsky_NIPS12}
A.~Krizhevsky, I.~Sutskever, and G.~E. Hinton.
\newblock Image{N}et classification with {D}eep {C}onvolutional {N}eural
  {N}etworks.
\newblock In \emph{Conference on Neural Information Processing Systems (NIPS)},
  2012.

\bibitem[Lai et~al.(2011)Lai, Bo, Ren, and Fox]{Lai_ICRA11}
K.~Lai, L.~Bo, X.~Ren, and D.~Fox.
\newblock A large-scale hierarchical multi-view {RGB-D} object dataset.
\newblock In \emph{{IEEE} International Conference on Robotics and Automation
  (ICRA)}, 2011.

\bibitem[Lai et~al.(2014)Lai, Bo, and Fox]{Lai_icra14}
K.~Lai, L.~Bo, and D.~Fox.
\newblock Unsupervised feature learning for 3{D} scene labeling.
\newblock In \emph{IEEE International Conference on on Robotics and Automation
  (ICRA)}, 2014.

\bibitem[Liu et~al.(2016)Liu, Anguelov, D.~Erhan, Reed, Fu, and
  Berg]{Liu_ECCV16}
W.~Liu, D.~Anguelov, C.~Szegedy D.~Erhan, S.~Reed, C.~Fu, and A.~C. Berg.
\newblock {SSD}: Single shot multibox detector.
\newblock In \emph{European Conference on Computer Vision (ECCV)}, 2016.

\bibitem[Mahler et~al.(2016)Mahler, Pokorny, Hou, Roderick, Laskey, Aubry,
  Kohlhoff, Kroeger, Kuffner, and Goldberg]{DexNet_ICRA16}
J.~Mahler, F.~T. Pokorny, B.~Hou, M.~Roderick, M.~Laskey, M.~Aubry,
  K.~Kohlhoff, T.~Kroeger, J.~Kuffner, and Ken Goldberg.
\newblock Dex-net 1.0: A cloud-based network of 3{D} objects for robust grasp
  planning using a multi-armed bandit model with correlated rewards.
\newblock In \emph{International Conference on Robotics and Automation (ICRA)},
  2016.

\bibitem[Mousavian et~al.(2016)Mousavian, Pirsiavash, and
  Kosecka]{Arsalan_3DV16}
A.~Mousavian, H.~Pirsiavash, and J.~Kosecka.
\newblock Joint semantic segmentation and depth estimation with deep
  convolutional networks.
\newblock In \emph{IEEE International Conference on 3DVision (3DV)}, 2016.

\bibitem[Peng et~al.(2015)Peng, Sun, Ali, and Saenko]{SaenkoICCV15}
X.~Peng, B.~Sun, K.~Ali, and K.~Saenko.
\newblock Learning deep object detectors from 3{D} models.
\newblock In \emph{IEEE International Conference on Computer Vision (ICCV)},
  2015.

\bibitem[Ren et~al.(2015)Ren, He, Girshick, and Sun]{Ren_NIPS15}
S.~Ren, K.~He, R.~Girshick, and J.~Sun.
\newblock Faster {R-CNN}: Towards real-time object detection with region
  proposal networks.
\newblock In \emph{Conference on Neural Information Processing Systems (NIPS)},
  2015.

\bibitem[Richter et~al.(2016)Richter, Vineet, Roth, and Koltun]{KoltunECCV16}
S.~Richter, V.~Vineet, S.~Roth, and V.~Koltun.
\newblock Playing for data: Ground truth from computer games.
\newblock In \emph{European Conference on Computer Vision (ECCV)}, 2016.

\bibitem[Sermanet et~al.(2013)Sermanet, Eigen, Zhang, Mathieu, Fergus, and
  LeCun]{SermanetCVPR13}
P.~Sermanet, D.~Eigen, X.~Zhang, M.~Mathieu, R.~Fergus, and Y.~LeCun.
\newblock Overfeat: Integrated recognition, localization and detection using
  convolutional networks.
\newblock In \emph{IEEE Conference on Computer Vision and Pattern Recognition
  (CVPR)}, 2013.

\bibitem[Silberman et~al.(2012)Silberman, Hoiem, Kohli, and
  Fergus]{Silberman_ECCV12}
N.~Silberman, D.~Hoiem, P.~Kohli, and R.~Fergus.
\newblock Indoor segmentation and support inference from {RGB-D} images.
\newblock In \emph{European Conference on Computer Vision (ECCV)}, 2012.

\bibitem[Simonyan and Zisserman(2014)]{Simonyan_arXiv14}
K.~Simonyan and A.~Zisserman.
\newblock Very deep convolutional networks for large-scale image recognition.
\newblock \emph{CoRR}, abs/1409.1556, 2014.

\bibitem[Singh et~al.(2014)Singh, Sha, Narayan, Achim, and
  Abbeel]{Singh_ICRA14}
A.~Singh, J.~Sha, K.~Narayan, T.~Achim, and P.~Abbeel.
\newblock A large-scale 3{D} database of object instances.
\newblock In \emph{{IEEE} International Conference on Robotics and Automation
  (ICRA)}, 2014.

\bibitem[Song et~al.(2015)Song, Zhang, and Xiao]{Xiao_Robot}
S.~Song, L.~Zhang, and J.~Xiao.
\newblock Robot in a room: Toward perfect object recognition in closed
  environments.
\newblock In \emph{arXiv:1507.02703 [cs.CV] 9 Jul 2015}, July 2015.

\bibitem[Su et~al.(2015)Su, Qi, Li, and Guibas]{Su_ICCV15}
H.~Su, C.~Qi, Y.~Li, and L.~Guibas.
\newblock Render for cnn: Viewpoint estimation in images using cnns trained
  with rendered 3{D} model views.
\newblock In \emph{The IEEE International Conference on Computer Vision
  (ICCV)}, December 2015.

\bibitem[Sun et~al.(2016)Sun, Bianchi, Bohg, and Dollar]{Sun_WICRA16}
Y.~Sun, M.~Bianchi, J.~Bohg, and A.~Dollar.
\newblock http://rhgm.org/activities/workshopicra16/.
\newblock In \emph{Workshop on Grasping and Manipulation Datasets (ICRA)},
  2016.

\bibitem[Tanaka et~al.(2012)Tanaka, Kamio, and Okutomi]{Tanaka_SIGGRAPH12}
M.~Tanaka, R.~Kamio, and M.~Okutomi.
\newblock Seamless image cloning by a closed form solution of a modified
  poisson problem.
\newblock In \emph{SIGGRAPH Asia 2012 Posters}, SA '12, 2012.

\bibitem[Tang et~al.(2012)Tang, Miller, Singh, and Abbeel]{Tang_ICRA12}
J.~Tang, S.~Miller, A.~Singh, and P.~Abbeel.
\newblock A textured object recognition pipeline for color and depth image
  data.
\newblock In \emph{{IEEE} International Conference on Robotics and Automation
  (ICRA)}, 2012.

\bibitem[Taylor and Cowley(2012)]{Taylor_RSS12}
C.~Taylor and A.~Cowley.
\newblock Parsing indoor scenes using {RGB-D} imagery.
\newblock In \emph{Robotics: Science and Systems (RSS)}, July 2012.

\bibitem[Uijlings et~al.(2013)Uijlings, van~de Sande, Gevers, and
  Smeulders]{Uijlings_IJCV13}
J.~R.~R. Uijlings, K.~E.~A. van~de Sande, T.~Gevers, and A.~W.~M. Smeulders.
\newblock Selective search for object recognition.
\newblock \emph{International Journal of Computer Vision (IJCV)}, 2013.
\newblock \doi{10.1007/s11263-013-0620-5}.
\newblock URL
  \url{http://www.huppelen.nl/publications/selectiveSearchDraft.pdf}.

\end{thebibliography}

\end{document}